\documentclass[runningheads]{llncs}
\usepackage{amsmath,amsfonts,bm}

\def\eqref#1{equation~\ref{#1}}

\def\1{\bm{1}}

\def\eps{{\epsilon}}

\def\rva{{\mathbf{a}}}

\def\rvo{{\mathbf{o}}}

\def\rvw{{\mathbf{w}}}
\def\rvx{{\mathbf{x}}}
\def\rvy{{\mathbf{y}}}
\def\rvz{{\mathbf{z}}}

\def\rmI{{\mathbf{I}}}

\def\vzero{{\bm{0}}}

\def\va{{\bm{a}}}

\def\vf{{\bm{f}}}
\def\vg{{\bm{g}}}

\def\vo{{\bm{o}}}

\def\vq{{\bm{q}}}

\def\vw{{\bm{w}}}
\def\vx{{\bm{x}}}
\def\vy{{\bm{y}}}

\def\mF{{\bm{F}}}

\def\mI{{\bm{I}}}

\DeclareMathAlphabet{\mathsfit}{\encodingdefault}{\sfdefault}{m}{sl}
\SetMathAlphabet{\mathsfit}{bold}{\encodingdefault}{\sfdefault}{bx}{n}

\def\gD{{\mathcal{D}}}

\def\gK{{\mathcal{K}}}
\def\gL{{\mathcal{L}}}

\def\gN{{\mathcal{N}}}

\def\gT{{\mathcal{T}}}
\def\gU{{\mathcal{U}}}

\newcommand{\E}{\mathbb{E}}

\newcommand*{\dif}{\mathop{}\!\mathrm{d}}

\usepackage[utf8]{inputenc} 
\usepackage[T1]{fontenc}    
\usepackage{url}            
\usepackage{booktabs}       
\usepackage{nicefrac}       
\usepackage{microtype}      
\usepackage{graphicx}
\usepackage{float}
\usepackage{subfig}
\usepackage{mathtools}
\usepackage{svg}
\usepackage{wrapfig}
\usepackage{booktabs}
\usepackage{makecell}
\usepackage{array}
\usepackage{longtable}
\usepackage{tabularx}
\usepackage{caption}
\usepackage[hidelinks]{hyperref}
\usepackage{pifont}
\usepackage{textcomp}
\usepackage{lipsum}
\usepackage{wrapfig}
\usepackage{threeparttable}
\usepackage{cleveref}
\usepackage{stmaryrd}
\usepackage{algorithm}
\usepackage{algorithmic}
\usepackage{ulem}
\usepackage{multirow,tabularx}
\usepackage{enumitem}
\usepackage[square,sort,comma,numbers]{natbib}
\usepackage[title]{appendix}

\begin{document}
\title{BiKC: Keypose-Conditioned Consistency Policy for Bimanual Robotic Manipulation}
%
%
\newcommand{\repeatthanks}{\textsuperscript{\thefootnote}}

\author{
    Dongjie Yu\inst{1,}\thanks{Equal contribution} \and
    Hang Xu\inst{2,}\repeatthanks \and
    Yizhou Chen\inst{2} \and
    Yi Ren\inst{3} \and
    Jia Pan\inst{1,2,}\thanks{Corresponding author}
}
\authorrunning{D. Yu et al.}
\titlerunning{BiKC}
%
\institute{
    Department of Computer Science, The University of Hong Kong \\ \email{djyu@connect.hku.hk, jpan@cs.hku.hk} \and
    Centre for Transformative Garment Production, HKU \\ \email{\{hang.xu,yz.chen\}@transgp.hk} \and
    Advanced Manufacturing Lab, Huawei Technologies \\ \email{even.renyi@huawei.com}
}
\maketitle              
\begin{abstract}
Bimanual manipulation tasks typically involve multiple stages which require efficient interactions between two arms, posing step-wise and stage-wise challenges for imitation learning systems. 
Specifically, failure and delay of one step will broadcast through time, hinder success and efficiency of each sub-stage task, and thereby overall task performance.
Although recent works have made strides in addressing certain challenges, few approaches explicitly consider the multi-stage nature of bimanual tasks while simultaneously emphasizing the importance of inference speed.
In this paper, we introduce a novel keypose-conditioned consistency policy tailored for bimanual manipulation.
It is a hierarchical imitation learning framework that consists of a high-level keypose predictor and a low-level trajectory generator.
The predicted keyposes provide guidance for trajectory generation and also mark the completion of one sub-stage task.
The trajectory generator is designed as a consistency model trained from scratch without distillation, which generates action sequences conditioning on current observations and predicted keyposes with fast inference speed.
Simulated and real-world experimental results demonstrate that the proposed approach surpasses baseline methods in terms of success rate and operational efficiency. Codes are available at \href{https://github.com/ManUtdMoon/BiKC}{https://github.com/ManUtdMoon/BiKC}.

\end{abstract}
\section{Introduction}
\label{sec:introduction}

Bimanual manipulation are ubiquitous in daily activities and industrial applications such as cloth folding, battery slotting and tool assembling \citep{chi2024universal,zhao2023learning, grotz2024peract2, bahety2024screwmimic}.
These tasks typically require intricate coordination between two arms and involve multiple stages, resulting in more complexity than unimanual case ~\citep{gao2024bikvil, varley2024embodied}.
One intuitive solution is to assign a specific role to each arm based on human-specified rules, with each arm being programmed independently~\citep{krebs2022bimanual, grannen2023stabilize, liu2024voxactb}.
However, this approach proves to be labor-intensive and hard-coded, as coordination rules often require customization for specific tasks, particularly those involving multiple stages~\citep{liu2022robot}.
Recently, imitation learning (IL), \textit{a.k.a.} learning from demonstration (LfD), has been widely applied in bimanual manipulation. IL is a straightforward yet effective approach that obtains automated agents in an end-to-end manner by imitating demonstrations, thereby eliminating the need for explicitly and manually designing coordination rules~\citep{ding2019goalconditioned, xie2020deep,kujanpaa2023hierarchical}.

Nevertheless, applying IL trivially to bimanual manipulation faces significant challenges due to the multi-stage nature: multi-modal demonstrations, per-step efficiency and per-stage reliability.
First, demonstrations of bimanual cases tend to be diverse due to expanded degrees of freedom (DOF)~\cite{krebs2022bimanual}, resulting in multi-modal demonstrations in terms of behavioral styles or stage orders, making them challenging to imitate.
Second, multi-stage tasks generally have extended time duration, causing latency at each step to accumulate, resulting in prolonged operational times. This obviously harms efficiency, especially in real-world systems.
Third, the completion (i.e., success or failure) of each stage has a lasting impact on subsequent stages. Taking transferring cube as an example, if the robot fails to pick up the cube, it will certainly fail to hand it over.
Since failure may occur at any stage, multi-stage tasks are more prone to fail compared to single-stage ones.
Consequently, addressing multi-stage bimanual manipulation tasks is a challenging but crucial topic.

Recently, advancements have been made in bimanual IL, while few studies have comprehensively addressed these challenges.
A novel IL algorithm termed Action Chunking with Transformers (ACT) is proposed in~\citep{zhao2023learning} to mitigate compounding errors in multi-stage bimanual tasks. However, it disregards inevitable sub-goals leading to the final completion, resulting in decreased overall success.
Additionally, ACT fixes its behavior style during evaluation, lacking the ability to reproduce multi-modal distributions present in demonstrations.
Another inspiring IL method, Diffusion Policy (DP)~\citep{chi2023diffusion}, inherently possesses the capability to represent multi-modal distributions and can be naturally extended to bimanual cases. However, its inference latency poses a key limitation on efficient completion in real-world and multi-stage tasks.
Furthermore, while some works have recognized the issue of reliability in multi-stage tasks, they have solely focused on unimanual tasks.
Representative works assume that some discrete \textit{keyposes} (also called keyframes, next best poses) exist in continuous demonstrations, which serve as sub-goals and indicate the skeleton of the whole task~\citep{gervet2023act3d, xian2023chaineddiffuser,ma2024hierarchical}. By predicting the next keypose and connecting keyposes with continuous motion trajectories, these methods have achieved success in various multi-stage tasks. Although the keypose-based methods have demonstrated effectiveness, they have not yet been tailored and evaluated for bimanual manipulation.

In view of the above, we propose \underline{Bi}manual \underline{K}eypose-conditioned \underline{C}onsistency Policy (BiKC), a hierarchical IL framework that 1) learns from distributionally multi-modal demonstrations and 2) generates actions with fast inference for 3) multi-stage bimanual manipulation.
BiKC comprises a high-level \textit{keypose predictor} as a sub-goal planner and a low-level \textit{trajectory generator} as a behavior policy.
At the high level, the keypose predictor forecasts the next target keypose in joint space. The keypose acts as both an indicator for sub-stage completion and the guidance for low-level action generation, which enhances per-stage reliability and thereby improves overall success rate.
The low-level trajectory generator is formulated as a Consistency Model (CM) and trained from scratch through consistency training \citep{song2023consistency, song2024improved}. It generates a short-horizon action sequence conditioned on observations including vision and proprioception, as well as the target keypose provided by the keypose predictor. The action generation is akin to DP, yet it requires only one-step inference while maintaining sample quality.

We test BiKC on both simulated and real-world experiments built on the ALOHA platform, demonstrating that it improves overall success rates of multi-stage tasks with enhanced time efficiency during online execution. We further investigate BiKC's capability in time-sensitive dynamic tasks, such as place-and-pick on a conveyor belt, to highlight its advantages in inference speed.
\section{Related Work}
\label{sec:related_work}

\noindent \textbf{Imitation Learning for Robotic Manipulation.}
IL enables robots to acquire skills by learning directly from experts' demonstrations. The intuitive process makes it widely used in robotic manipulation.
IL in manipulation can be categorized into non-parametric approaches (e.g., Dynamic Movement Primitives~\citep{ijspeert2013dynamical}) and parametric methods promising for high-dimensional tasks~\citep{xie2020deep,sundaresan2024rtsketch}.
Behavioral cloning (BC)~\citep{florence2021implicit} is a classic parametric IL algorithm that acts similarly as supervised learning, directly fitting the mapping from observations (inputs) to actions (labels). 
Despite its simplicity and efficacy, BC-based approaches are susceptible to compounding errors~\citep{zhao2023learning} and multi-modality in demonstrations~\citep{chi2023diffusion}, which is worse when it comes to bimanual cases.
Recently, the integration of generative modeling techniques into IL has brought success~\citep{chi2024universal,zhao2023learning,chi2023diffusion, shi2023waypointbased}.
Predicting action sequences in these works mitigates compounding errors by shortening the effective horizon~\citep{zhao2023learning, chi2023diffusion, shi2023waypointbased}.
Some works \citep{zhao2023learning, shi2023waypointbased} address the multi-modality challenge by leveraging Variational Auto-Encoder (VAE)~\citep{kingma2014autoencoding} to focus on the predominant behavior style. Conversely, other works capture multi-modal action distributions by modeling the policy as a diffusion model~\citep{chi2023diffusion,ho2020denoising,song2020scorebased}.
However, diffusion models inherently incur significant inference latency due to iterative denoising, which presents a critical concern especially in real-world deployment.
In this work, we focus on a variant of diffusion models, Consistency Models (CMs)~\citep{song2023consistency, song2024improved}, to formulate robot policies.
CMs can capture multi-modal demonstrations like DP~\citep{song2023consistency} while generating action sequences with only one-step inference, i.e., fast inference speed.
Some concurrent works~\citep{lu2024manicm, prasad2024consistency} also apply CM to robotic manipulation tasks with the aim of accelerating run-time inference. However, these approaches distill CMs from well-trained diffusion models (i.e., consistency distillation (CD) \citep{song2023consistency}) and focus only on unimanual tasks. 
Here, CD necessitates additional efforts to learn a diffusion model first and depends on the quality of the learned diffusion model. To this end, we adopt consistency training (CT) \cite{song2023consistency, song2024improved} in this paper.
Through comprehensive experiments, we demonstrate that CMs \textit{trained from scratch} can achieve high-performance deployment and real-time reaction in bimanual tasks.

\noindent \textbf{Hierarchical IL for Multi-stage Tasks.} 
Hierarchical IL (HIL) is proposed to enhance the performance for multi-stage tasks, by decomposing a complex task into smaller and more manageable ones.
One line of studies pre-define a set of manipulation primitives, such as picking, placing, moving and stabilizing. Then a high-level policy is learned to sequentially choose primitives to be executed~\citep{luo2023multistage}.
In bimanual cases, this intuition is implemented by assigning a fixed role for each arm, after which each arm learns an independent policy for its designated role~\cite{grannen2023stabilize}. In this way, the bimanual system is decomposed into two unimanual cases. However, such a method is constrained to specific types of tasks.
Another line of work segments the demonstration into sub-stage trajectories. For instance, \citep{zhu2022bottomup} employs a bottom-up clustering on observations to accomplish unsupervised segmentation, while \citep{ma2024hierarchical} establishes heuristic rules based on robot kinematics.
One technique to segment trajectories is extracting \textit{keyposes}.
Keypose typically refers to the next-best end-effector (EE) pose or joint positions where the arm reaches a specific state, e.g., EE velocity approaching zero or the gripper beginning to open or close.
With identified keyposes, IL on a multi-stage task comes down to modeling both the high-level keypose transition and the low-level sub-trajectories connecting two-keypose pairs. Built on this idea, several works have made achievements in unimanual manipulation~\citep{gervet2023act3d,xian2023chaineddiffuser, ma2024hierarchical}. 
However, applying keypose-based approaches in bimanual tasks poses a challenge in coordinating two arms. In this study, we address this issue by adopting a merging method.
Furthermore, we condition the learning of the trajectory generator on target keyposes, to improve per-stage reliability and enhance overall success rates.

\section{Preliminaries: Consistency Models as Visuomotor Policy}
\label{sec:preliminary}

In this section, we first briefly introduce consistency models (CMs) as a generative modeling approach. Then we show the fundamental \textit{one-step generation} property of CMs can be leveraged in IL settings as visuomotor robot policies, compared with their SDE (stochastic differential equation) counterparts, Denoising Diffusion Probabilistic Models (DDPMs)~\citep{chi2023diffusion, ho2020denoising}.

\noindent\textbf{Consistency Models in Generative Modeling.}
CMs are proposed to remove the iterative sampling process in DDPMs to accelerate generation speed~\citep{song2023consistency} as another score-based generative modeling method. The core of CMs is the existence of an ordinary differential equation (ODE) shares the same per-step distribution $p_\sigma(\rvx)$ of samples $\rvx^\sigma$ with diffusion SDE~\cite{song2020scorebased}:
\begin{equation}
\label{eq:sde_to_ode}
\text{SDE:}
    \dif\rvx = \sqrt{2\sigma}\dif\rvw^\sigma
    \longleftrightarrow
\text{ODE:}
    \dif\rvx =\left[ -\sigma\nabla \log p_\sigma(\rvx) \right]\dif\sigma, \quad \sigma \in [\epsilon, \sigma_\text{max}],
\end{equation}
where $\sigma$ is the continuous noise level, $\{\rvw^\sigma\}_{\sigma\in[\epsilon,\sigma_\text{max}]}$ denotes the standard Brownian motion, $\nabla\log{p_\sigma(\rvx)}$ is the score function of $p_\sigma(\rvx)=\int p_\text{data}(\rvy)\gN(\rvx|\rvy,\sigma^2 \mI)\dif\rvy$. $\epsilon$ is a fixed small positive number such that $p_{\epsilon}(\rvx) \approx p_{\text{data}}(\rvx)$ and $\sigma_\text{max}$ is sufficiently large such that $p_{\sigma_\text{max}}(\rvx) \approx \gN(0,\sigma_\text{max}^2 \mI)$. Without perturbations from random noise at each diffusion step in DDPM, intermediate samples $\rvx^\sigma$ in the ODE are determined once the original data $\rvx$ is given. In other words, all samples along the same solution trajectory of the right part in \cref{eq:sde_to_ode} correspond to the same origin, which is called \textit{self-consistency condition} in~\citep{song2023consistency}:
\begin{equation}
    \vf(\rvx^{\sigma_\text{max}}, \sigma_\text{max}) = \vf(\rvx^{\sigma}, \sigma) = \cdots = \vf(\rvx^{\epsilon}, \epsilon) = \rvx^\epsilon, \quad \forall \sigma \in [\epsilon, \sigma_\text{max}],
    \label{eq:self_consistency}
\end{equation}
where $\vf(\cdot,\cdot)$ is the consistency function mapping diffused samples to origin.

\noindent\textbf{Training and Sampling of Consistency Models.}
Training CMs is to approximate the consistency function $\vf$ with a parameterized one $\vf_{\bm{\theta}}$ (typically neural networks) by learning to enforce \cref{eq:self_consistency}. Recall that $\rvx^\sigma$ is subject to $p_\text{data}(\rvx)*\gN(0,\sigma^2 \mI)$. Hence, two adjacent discretized samples on the same ODE trajectory can be represented as $\rvx+\sigma_n\rvz$ and $\rvx+\sigma_{n+1}\rvz$ where $\rvx\sim p_\text{data}$ and $\rvz\sim \gN(\bm{0},\mI)$. We can apply the self-consistency condition by minimizing the error between outputs of two adjacent samples. Then, the training objective of CMs is
\begin{equation}
\label{eq:cm_loss}
    \min_{\bm{\theta}} \gL({\bm{\theta}}, {\bm{\theta}}^{-}) = \E \lbrack \lambda(\sigma_n) d(\vf_{\bm{\theta}}(\rvx+\sigma_{n+1}\rvz, \sigma_{n+1}), \vf_{\bm{\theta}^-}(\rvx + \sigma_n \rvz, \sigma_n)) \rbrack,
\end{equation}
where $\lambda$ is a weighting function, $d$ is a distance metric, $n$ is the $N$-discretized time step ($\eps=\sigma_0<\cdots<\sigma_n<\cdots<\sigma_N=\sigma_\text{max}$) and $\bm{\theta}^-$ is the exponential moving average (EMA) of $\bm{\theta}$, \textit{a.k.a.} the target network. More details regarding design decisions for CMs are provided in Appendix A.
With a well-trained CM $\vf_{\bm{\theta}}$, we can draw samples from $p_\text{data}$ by sampling from the tractable Gaussian distribution $\hat{\rvx}^{\sigma_\text{max}}\sim\gN(\bm{0},{\sigma_\text{max}}^2\mI)$ and performing one-step inference $\hat{\rvx}^\eps=\vf_{\bm{\theta}}(\hat{\rvx}^{\sigma_\text{max}},{\sigma_\text{max}})$.

\noindent\textbf{Consistency Models as Visuomotor Policies.}
By analogy to Diffusion Policy~\citep{chi2023diffusion}, CMs can also be adapted to IL settings, serving as robot visuomotor policies. 
In this case, a CM characterizes the conditional distribution $p(\rva_t|\rvo_t)$ of robot action sequences $\rva_t$ given observation $\rvo_t$ from demonstrations.
At $t$-th step of closed-loop control, a consistency policy $\bm{\pi}$ predicts an action sequence by evaluating $\rva_t=\bm{\pi}_{\bm{\theta}}(\hat{\rva_t}^{\sigma_\text{max}},{\sigma_\text{max}}\mid \rvo_t)$ once. 
The key distinguishing feature of consistency policies, compared to diffusion policies, is their ability to generate an action sequence through only one-step inference without iterative sampling, thereby being faster while maintaining modeling quality.
This makes consistency policies computationally efficient during execution, enabling them to complete dynamic tasks besides quasi-static ones.

\section{Bimanual Keypose-Conditioned Consistency Policy}
\label{sec:method}

\begin{figure}[!t]
    \centering
    \includegraphics[width=0.85\textwidth]{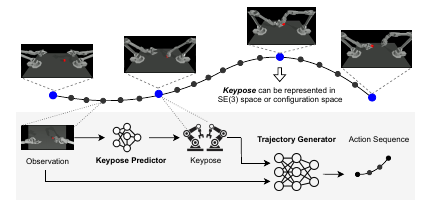}
    \vspace{-4mm}
    \caption{Illustration of the keypose-conditioned workflow.}
    \label{fig:overview}
    \vspace{-6mm}
\end{figure}

The proposed hierarchical bimanual manipulation framework BiKC is composed of a high-level keypose predictor and a low-level trajectory generator.
Here \textit{keyposes}~\citep{ijspeert2013dynamical} refer to inevitable commonalities across different demonstration trajectories of a specific task, either in the form of end-effector SE(3) pose or joint positions.
Each keypose indicates the effect of a subtask~\citep{garrett2021integrated} and the pre-condition for the following subtask, while a sequence of keyposes outlines the skeleton of a multi-stage task. 
As shown in the top half of Figure~\ref{fig:overview}, keyposes during transferring a cube include grasping the cube by right arm, passing it to the left.
The bottom half of Figure \ref{fig:overview} depicts online workflow of BiKC: 1) the keypose predictor updates target keypose given current observation and last keypose, when the deviation between the current state and the target keypose goes below a threshold; 2) the trajectory generator determines a short-horizon action sequence given both the historical observations and the target keypose.
Through keypose guidance, the trajectory generator has enhanced awareness of the critical junctures involved in completing each stage of a task, which is empirically claimed in Section~\ref{sec:experiment}.

\noindent\textbf{Problem Formulation.}
We assume a set of demonstrated trajectories $\gD = \{\tau^i \}_{i=0}^{D-1}$ collected via teleoperation~\citep{zhao2023learning}, where
$\tau^i$ is the $i$-th sequence of observation-action pairs, i.e., $\tau^i = \{ (\vo_t^i, \va_t^i) \}_{t=0}^{T_i-1}$. 
The observation $\vo_t^i = (\vx_t^i, \vq_t^i) $ includes RGB images $\vx_t^i$ and proprioceptive information $\vq_t^i$ (i.e., joint positions and gripper width), while the action $\va_t^i$ is the expected proprioceptive status. Note that we omit the superscript $i$ for simplicity if the context is clearly within one trajectory.
In any trajectory $\tau^i$, we assume a set of keyposes $\gK^i=\{k_0^i, \dots, k^i_j, \dots,k_{M_i}^i\}$ exists, which can be represented in either SE(3) or the configuration space. We choose the latter one in this work (i.e., joint positions of bimanual arms) to align with the representation of states and actions.
With $\gD$ and $\gK$, we aim to learn a keypose predictor $\vg_{\bm{w}}(\vo, k)$ for high-level guidance and a trajectory generator $\bm{\pi}_{\bm{\theta}}(\rva | \rvo, k)$ producing action sequences $\rva$ for task fulfillment conditioned on historical observations $\rvo$ and target keypose $k$.
We introduce design details of the keypose predictor and the trajectory generator below.

\subsection{Keypose Predictor}
Although the notion of a keypose predictor $\bm{g}$ has been proposed for unimanual manipulation~\citep{gervet2023act3d, xian2023chaineddiffuser, ma2024hierarchical}, its application to bimanual cases remains unresolved. This is due to the intricate coordination between the two arms~\citep{krebs2022bimanual}, which renders identifying bimanual keyposes challenging.
We propose a simple yet effective holistic manner to mitigate this issue and further train the keypose predictor.

\noindent\textbf{Bimanual Keypose Identification.}
We summarize and generalize the heuristics in previous studies~\citep{gervet2023act3d,xian2023chaineddiffuser,ma2024hierarchical} into three categories to decide whether a robotic arm is at a keypose: (1) changes in robot-object contact mode; (2) motion stalling and (3) case-specific rules. In more detail, changes in contact mode always come with the opening or closing of robotic grippers, which indicates changes in positional relationships among objects and changes in workspace state. Motion stalling means the arm is performing pre-grasp or pre-release, which are significant for fine-grained manipulation. Case-specific rules are dependent on spatial-temporal relationship between two arms in certain tasks, e.g., the relative distance between left and right grippers and the height of grippers w.r.t. the table-top. Specifically, we scan each demonstration trajectory and check whether (1) the gripper begins to open or close; (2) the velocity of gripper is below a threshold; and (3) the relative distance or height is below a threshold and mark those timesteps that satisfy the conditions as keyposes.

The extension from unimanual to bimanual is non-trivial due to the coordination between two arms to accomplish a shared objective. In unimanual cases, the arm is free to move on once it reaches the current target.
However, bimanual scenarios involve certain keyposes that necessitate interaction and synchronization between the two arms.
Consider the third keypose of cube transfer in Figure~\ref{fig:overview} as an illustration: the left arm must await the right one to finish the handover. If the left arm immediately advances towards the subsequent keypose upon reaching the handover keypose, it will fail to grasp the cube. Previous work summarizes the taxonomy of bimanual coordination comprehensively~\citep{krebs2022bimanual}.
Although it is possible to use the taxonomy to decide the legal behavior (pause or proceed) of both arms at a certain keypose, applying this directly to keypose identification and prediction requires additional manual annotation and interference.

\begin{figure}[!ht]
\vspace{-5mm}
    \centering
    \includegraphics[width=0.65\textwidth]{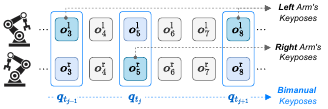}
    \vspace{-3mm}
    \caption{Extracting bimanual keyposes by merging ones of each arm.}
    \label{fig:bimanual_keyposes}
\vspace{-5mm}
\end{figure}

We propose a \textit{merging} method rather than explicitly modeling the complex relationships between the two arms' keyposes as shown in Figure~\ref{fig:bimanual_keyposes}. In other words, we simply identify keyposes for each arm, take the results of \textit{union} as keyposes for the \textit{bimanual} system.
Specifically, we assume that keypose indices of left arm are $t_1^\text{l}, t_2^\text{l},\cdots$ while those of right arm are $t_1^\text{r}, t_2^\text{r},\cdots$. Then the bimanual keypose indices can be obtained by
\begin{equation}
\texttt{sort}(\texttt{unique}(t_1^\text{l}, t_2^\text{l},\cdots, t_1^\text{r}, t_2^\text{r},\cdots)).   
\end{equation}
The rationality behind this is two-fold: (1) when the two arms are interacting with each other, they always share the same keypose simultaneously because of their spatial adjacency; (2) when the two arms are independent in spatial domains, and one arm reaches a keypose, it is acceptable to mark this timestep as a milestone keypose for the other arm, because the latter arm is progressing towards its own target keypose regardless. Merging keyposes naturally eliminates the need for manual coordination of pause-or-proceed decisions, allowing each arm to seamlessly transition to the subsequent keypose after reaching the current one, as synchronization has already been established.

\vspace{1mm}
\noindent \textbf{Training Keypose Predictor.}
We can extract bimanual keyposes $\gK^i=\{ k_j^i \}_{j=0}^{M_i}$ from each demonstration $\tau^i= \{ (\vo_t^i, \va_t^i) \}_{t=0}^{T_i-1}$. Here, $k_j^i$ is the $j$-th keypose in the $i$-th trajectory in the form of joint positions such that $k_j^i=\vq_{t_j}^i$, and $t_j$ is the corresponding timestep in the trajectory ($t_0=0, t_{M_i}=T_i$). Note that the number of keyposes is much smaller than the number of timesteps (i.e., $M_i\ll T_i$). Therefore, we enrich the training data for keypose predictor by training it to predict target keyposes at all timesteps, rather than only at the designated keypose steps. This approach enhances the predictor's capability by leveraging a larger volume of training data. Specifically, the keypose dataset is constructed as
\begin{equation}
    \label{eq:keypose_dataset}
    \gT_{\text{keypose}} = \bigcup\limits_{i=0}^{D-1} \bigcup\limits_{j=0}^{M_i-1} \bigcup\limits_{t=t_j}^{t_{j+1}-1} \{ (\vo^i_t, k^i_{j}, k^i_{j+1}) \},
\end{equation}
which is composed of tuples including the observation $\vo_t^i$ and keyposes preceding and succeeding it.
The keypose predictor $\hat{k}_{j+1}=\vg_{\bm{w}}(\vo_t, k_j)$ is learned by minimizing the Mean Square Error (MSE) between the predicted keyposes and the ground truth as follows:
\begin{equation}
\label{eq:keypose_loss}
    \gL_{\text{keypose}}(\vw) = \mathop{\E}\limits_{(\vo, k, k') \sim \gT_{\text{keypose}}} \Arrowvert \vg_{\bm{w}}(\vo, k) - k' \Arrowvert^2_2.
\end{equation}

\subsection{Trajectory Generator}

We formulate the trajectory generator, which plans short-horizon action sequences, as a Consistency Model (CM)~\cite{song2023consistency, song2024improved}. This leverages CM's capability of modeling complex distributions and one-step sampling to reduce inference latency.
Incorporating requirements outlined in Section~\ref{sec:preliminary} and \ref{sec:method}, this CM policy models action distribution conditioned on the target keypose and historical observations.

\begin{figure}[!t]
    \centering
    \includegraphics[width=0.78\textwidth]{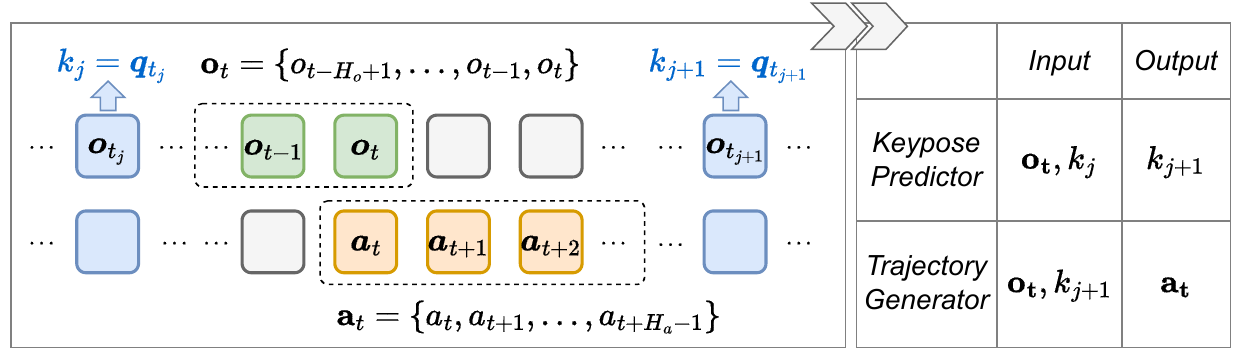}
    \vspace{-3mm}
    \caption{Extracting keypose and local trajectory samples from demonstrations.}
    \label{fig:traj_sample}
\vspace{-6mm}
\end{figure}

Specifically, the keypose-conditioned trajectory generator is trained on an action sequence dataset built on demonstrations $\gD = \{\tau^i\}_{i=0}^{D-1}$ and identified keyposes $\{ \gK^i \}_{i=0}^{D-1}$ as shown in Figure~\ref{fig:traj_sample}. The dataset can be represented as
\begin{equation}
    \label{eq:traj_dataset}
    \gT_{\text{traj}} = \bigcup\limits_{i=0}^{D-1} \bigcup\limits_{j=0}^{M_i-1} \bigcup\limits_{t=t_j}^{t_{j+1}-1} \{ ( \rvo^i_t, k^i_{j+1}, \rva^i_t) \},
\end{equation}
where $\rvo_t = \{ \vo_{t-H_{o}+1}, \dots, \vo_{t-1}, \vo_t \}, \rva_t = \{\va_t, \va_{t+1}, \dots, \va_{t+H_{a}-1} \}$, $H_o$ and $H_a$ denote the length of historical observations and action sequence, respectively.
The target keypose $k^i_{j+1} = \vq^i_{t_{j+1}}$ is joint positions at the next keypose timestep index exactly after $t$.
One special case is that if the indices of actions exceed the index of target keypose (i.e., $t+H_a-1 \ge t_{j+1}$), we pad actions after $t_{j+1}$ with $\va_{t_{j+1}-1}$ to keep the system staying at the keypose. 

The keypose-conditioned trajectory generator $\bm{\pi}_{\bm{\theta}}(\rva_t | \rvo_t, k_{j+1})$ is learned by minimizing the consistency training objective
\begin{equation}
\label{eq:traj_loss}
    \gL_{\text{traj}}(\bm{\theta}, \bm{\theta^{-}}) = \mathop{\E}\limits_{(\rvo, k, \rva) \sim \gT_{\text{traj}},n} \lbrack \lambda(\sigma_n) d(\bm{\pi}_{\bm{\theta}}(\rva+ \sigma_{n+1}\rvz, \sigma_{n+1} | \rvo, k), \bm{\pi}_{\theta^{-}}(\rva + \sigma_n \rvz, \sigma_n |\rvo, k)) \rbrack,
\end{equation}
where $\rvz \sim \gN(\vzero, \rmI)$, $\bm{\theta}^-\gets \text{stopgrad}(\mu\bm{\theta^-}+(1-\mu)\bm{\theta})$ with $\mu$ representing the EMA decay rate and $\text{stopgrad}$ meaning stopping gradient calculation. During training, the keypose fed into $\bm{\pi}_{\bm{\theta}}$ is the ground truth in the demonstration while during deployment, the keypose is predicted by the keypose predictor $\bm{g}_{\bm{w}}$.

\subsection{Implementation and Training Details}
\label{sec:method_details}

We train $\bm{g}_{\bm{w}}$ and $\bm{\pi}_{\bm{\theta}}$ separately and chain them together when deployment. The keypose predictor resembles a vision transformer. 
The keypose predictor $\vg_{\bm{w}}$ takes $640\times480$ RGB images as input. These images are processed by a non-pretrained ResNet18 encoder into a feature sequence. This feature sequence is then concatenated with the embedding of the last predicted keypose, forming a memory sequence. The query embedding attends to this memory sequence to output the next keypose prediction.

Design choices of $\bm{\pi}_{\bm{\theta}}$ generally follows those of CNN-based diffusion policy proposed in \citep{chi2023diffusion}. 
However, a significant modification we made is that the 1D temporal CNN backbone conditions on target keyposes besides visual and proprioceptive features.
We also use the same settings of $H_o=2$ and $H_a=8$. On CM training side, we leverage part of improved techniques proposed in \citep{song2024improved} to enhance experimental results. Specifically, the weighting function is $\lambda(\sigma_n) = 1/(\sigma_{n+1}-\sigma_n)$ while the distance metric is Pseudo-Huber loss $d(\vx,\vy)=\sqrt{\Vert \vx-\vy\Vert^2_2+c^2}-c$ with $c=0.0064$. Furthermore, we adopt $\mu=0$ and the step-shaped discretization curriculum. Both models are trained by AdamW~\citep{loshchilov2018decoupled} optimizer, with batch size $B=64$ and learning rate cosine annealing from 1e-4 to zero across all tasks.
Each element of inputs to both models are linearly normalized to $[-1,1]$. Then outputs will be unnormalized back to original ranges.
More detailed parameters can be found in Appendix B.
\section{Experiment}
\label{sec:experiment}


The proposed BiKC is evaluated on both simulated and real-world manipulation tasks in order to address the following research questions: 
1) Does the incorporation of keypose contribute to performance improvements in multi-stage bimanual manipulation tasks? 
2) Does CM-based policies reduce inference latency? %
3) Can BiKC enhance operational efficiency while also improve whole task success rate in real-world multi-stage bimanual tasks?

We compare BiKC with state-of-the-art (SOTA) imitation learning baselines.
\textbf{ACT} \citep{zhao2023learning} formulates the policy as a conditional variational autoencoder (cVAE), predicting a sequence of actions based on current observation. It adopts temporal ensemble to enhance action smoothness. However, ACT is designed to capture the primary behavior style, lacking the capability to handle multi-modality.
\textbf{DP} \citep{chi2023diffusion} learns the policy as a diffusion model~\cite{ho2020denoising, song2020scorebased}, which denoises an action sequence iteratively from random noise, conditioned on historical observations. The action generation requires iterative network forward pass, resulting in substantial inference latency.
Additionally, \textbf{CP} (Consistency Policy) is designed as a non-keypose version to serve as an ablation baseline. Here, the policy is formulated as a consistency model \citep{song2023consistency, song2024improved} and learned from scratch by consistency training loss in Equation~\ref{eq:cm_loss}. CP predicts an action sequence conditioned on historical observations, which can be achieved through a single-step inference.

\subsection{Bimanual Simulation}

\begin{figure}[!t]
\centering
\subfloat[]{\label{fig:sim_tasks}\includegraphics[width=0.78\textwidth]{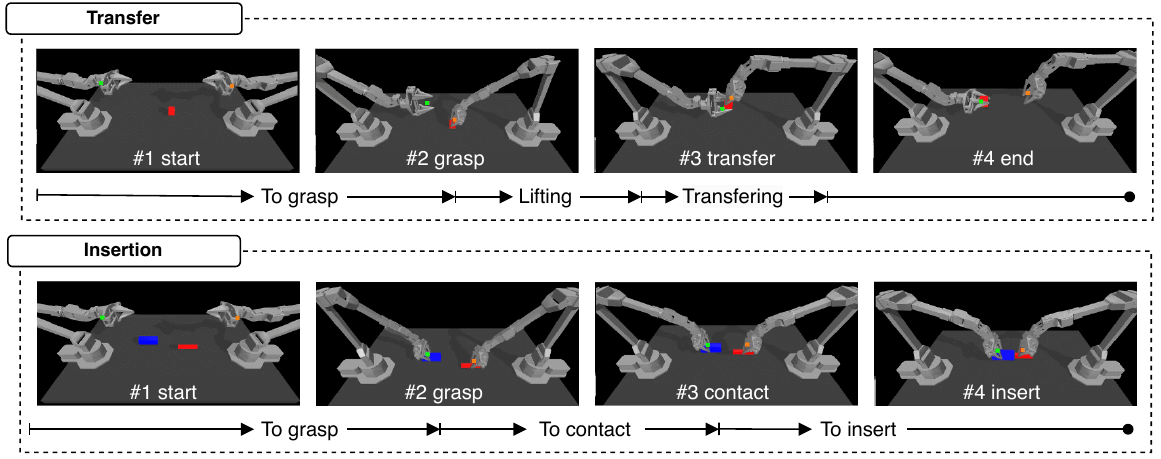}}
\subfloat[]{\label{fig:sim_obs}\includegraphics[width=0.19\textwidth]{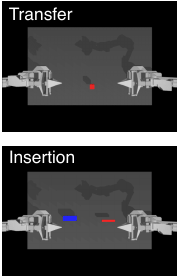}}
\vspace{-3mm}
\caption{
(a) Snapshots of sub-stages and predicted keyposes (green for left arm while orange for right arm).
(b) Rendered RGB observation.
}
\label{fig:sim_results}
\vspace{-2mm}
\end{figure}

\noindent\textbf{Tasks and Settings.}
We evaluate BiKC on two bimanual tasks based on MuJoCo~\citep{todorov2012mujoco}, \textbf{Transfer} and \textbf{Insertion}, developed by~\citep{zhao2023learning}.
Both tasks involves multiple stages and find-grained coordination between two arms, as illustrated in Figure \ref{fig:sim_tasks}.
For \textbf{Transfer} task, four keyposes (i.e., \textit{start}, \textit{grasp}, \textit{transfer}, \textit{end}) are extracted as boundaries of three sub-stages, including: 1) right arm moves to fetch the cube, 2) the cube is grasped and lifted, and 3) the cube is handed over from the right to the left. 
\textbf{Insertion} also involves three stages separated by four keyposes (i.e., \textit{start}, \textit{grasp}, \textit{contact}, \textit{insert}), including: both arms 1) grasp the peg and socket, 2) lift peg and socket to allow contact, and 3) finally finish the bimanual insertion.
The simulation environment includes two Interbotix vx300s robotic arm on both sides of a table-top. 
As shown in Figure~\ref{fig:sim_obs}, the observation $\vo_t$ consists of a $480\times640$ RGB image rendered by MuJoCo from a static top-down view, as well as a 14D vector of current joint positions for both arms.
The action is the expected joint positions of the two arms for the next timestep, which is also a 14D vector.
The initial location of the cube, the socket and the peg are all randomly generated within a rectangle area in both demonstrations and evaluation rollouts. We refer to \citep{zhao2023learning} for specific details.
We use the 50 demonstrations open-sourced by~\citep{zhao2023learning} to train all models for each task. They are collected using a scripted policy at a control frequency of 50Hz, with each trajectory lasting 400 timesteps (i.e., 8 seconds).
In the simulation experiments, we primarily focus on the impact of keypose on the success rate, while we disregard the evaluation of inference latency in the simulated tasks as the simulation is paused during policy inference.

\begin{table}[!t]
\centering
\vspace{-3mm}
\caption{Success rate (\%) of each sub-stage in two simulated tasks.}
\vspace{0.1cm}
\begin{tabular}{@{}clccccc@{}}
\toprule
\textbf{Task}             & \textbf{Sub-stage} & \textbf{ACT} & \textbf{ DP} & \textbf{\hspace{3mm} CP | {\textit{BiKC}}}\\ \midrule
\multirow{3}{*}{Transfer} & To grasp           & \textbf{97}  & \textbf{96}   & \textbf{96} | \textbf{96} \\
                          & Lifting            & 90           & \textbf{96}       & 93 | \textbf{95} \\
                          & Transferring       & 86           & \textbf{96}       & 93 | \textbf{95} \\
\midrule
\multirow{3}{*}{Insertion}& To grasp           & \textbf{93}     & 83             & 83 | \textbf{98} \\
                          & To contact         & \textbf{90}     & 47             & 53 | \textbf{96}  \\
                          & To insert          & 32              & 37             & \textbf{38} | \textbf{43}\\
\bottomrule
\end{tabular}
\label{tab:sim_results}
\vspace{-6mm}
\end{table}


\noindent\textbf{Results.}
We report the average success rate of each sub-stage in both simulated tasks in Table \ref{tab:sim_results}. All results are averaged across 3 seeds and 50 rollouts. Note that the sub-stages are defined same as~\cite{zhao2023learning} so we directly report results of ACT in the original paper.
The top-two results are highlighted, which are predominantly achieved by BiKC. 
Particularly, for the \textbf{Insertion} task, BiKC significantly outperforms CP, with improvements of 16\%, 43\%, and 5\% for each sub-task, respectively. 
%
Although ACT achieves highest success rates in some sub-tasks, its overall success rates for whole tasks are inferior. We suppose that it is because the third stage of both tasks involves fine-grained cooperation between two arms and is sensitive to operation accuracy. In other words, small errors will lead to collision and task failure.
With guidance from keyposes, BiKC is aware of the fine-grained sub-target and finally reaches higher success rates.

Additionally, we visualize the predicted keyposes during task execution in Figure~\ref{fig:sim_tasks}. The predicted keyposes serve as meaningful boundaries segmenting sub-tasks. Taking the \textbf{Transfer} task as an instance, we observe that keyposes are updated at the moments including starting step, the cube to be grasped, the position to handover and the end step. These sequential keyposes are common across all demonstrations and represent the skeleton of successfully completing the entire task, so it is reasonable to use these keyposes to guide action generation and identify the completion of sub-stages.

\subsection{Real-World Experiments}


%

We evaluate BiKC on three real-world tasks featuring multiple sub-stages, intricate coordination, and rich contact: Screwdriver Packing, Placing and Picking on Conveyor, and Cup Insertion.
%
The \textbf{Screwdriver Packing} task aims to demonstrate BiKC's ability to enhance sub-stage success rate while improving operational efficiency.
The \textbf{Placing and Picking on Conveyor} task features a dynamic object, allowing us to emphasize the advantage of CM over DP in terms of inference speed.
Finally, we assert that BiKC can capture multi-modality in \textbf{Cup Insertion} task, overcoming the limitations of ACT.

\begin{figure}[!hb]
\vspace{-3mm}
    \centering
    \includegraphics[width=0.75\textwidth]{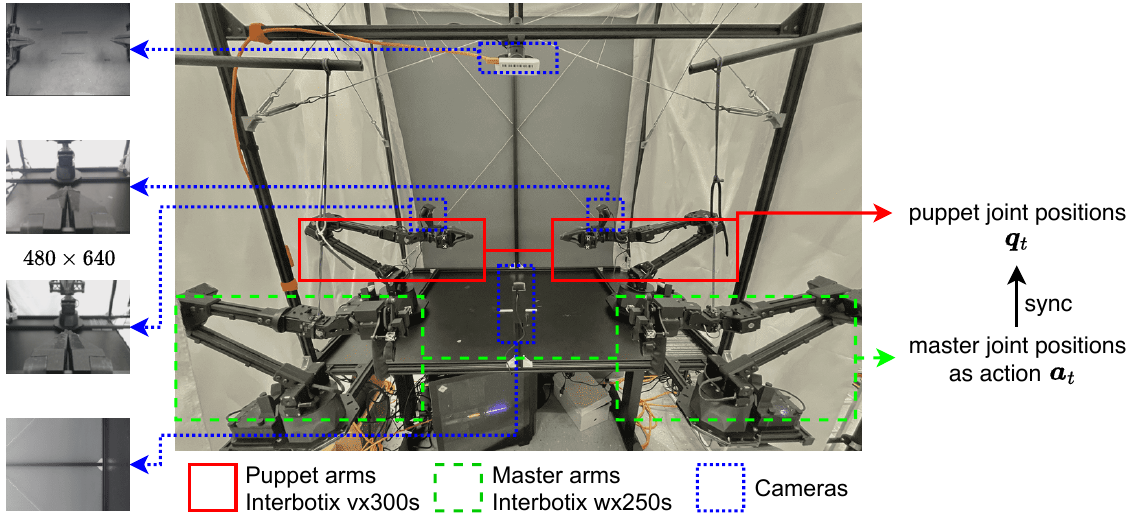}
    \vspace{-3mm}
    \caption{Real-world ALOHA platform.}
    \label{fig:real_platform}
\vspace{-5mm}
\end{figure}

\noindent\textbf{Robot Platform.}
We run real-world experiments with \underline{a} \underline{lo}w-cost open-source \underline{ha}rdware system for bimanual teleoperation (ALOHA)~\citep{zhao2023learning}.
ALOHA is comprised of two 7DOF master arms, two 7DOF puppet arms and accessories, as depicted in Figure~\ref{fig:real_platform}. 
During teleoperation for recording demonstrations, a human manipulates the master arms, with their joint positions stored as action $\va_t$ and transmitted to the puppet arms for synchronization and execution.
During deployment, the expected joint positions of puppet arms are provided by neural policies for autonomous execution.
Four cameras installed in ALOHA offers visual observations.
Two are positioned on the front and at the top of the frame, providing a static overall view of the workspace.
The other two are mounted on the wrists of the puppets, offering closed views for fine-grained manipulation.

\begin{figure}[!ht]
    \centering
    \vspace{-6mm}
    \includegraphics[width=\textwidth]{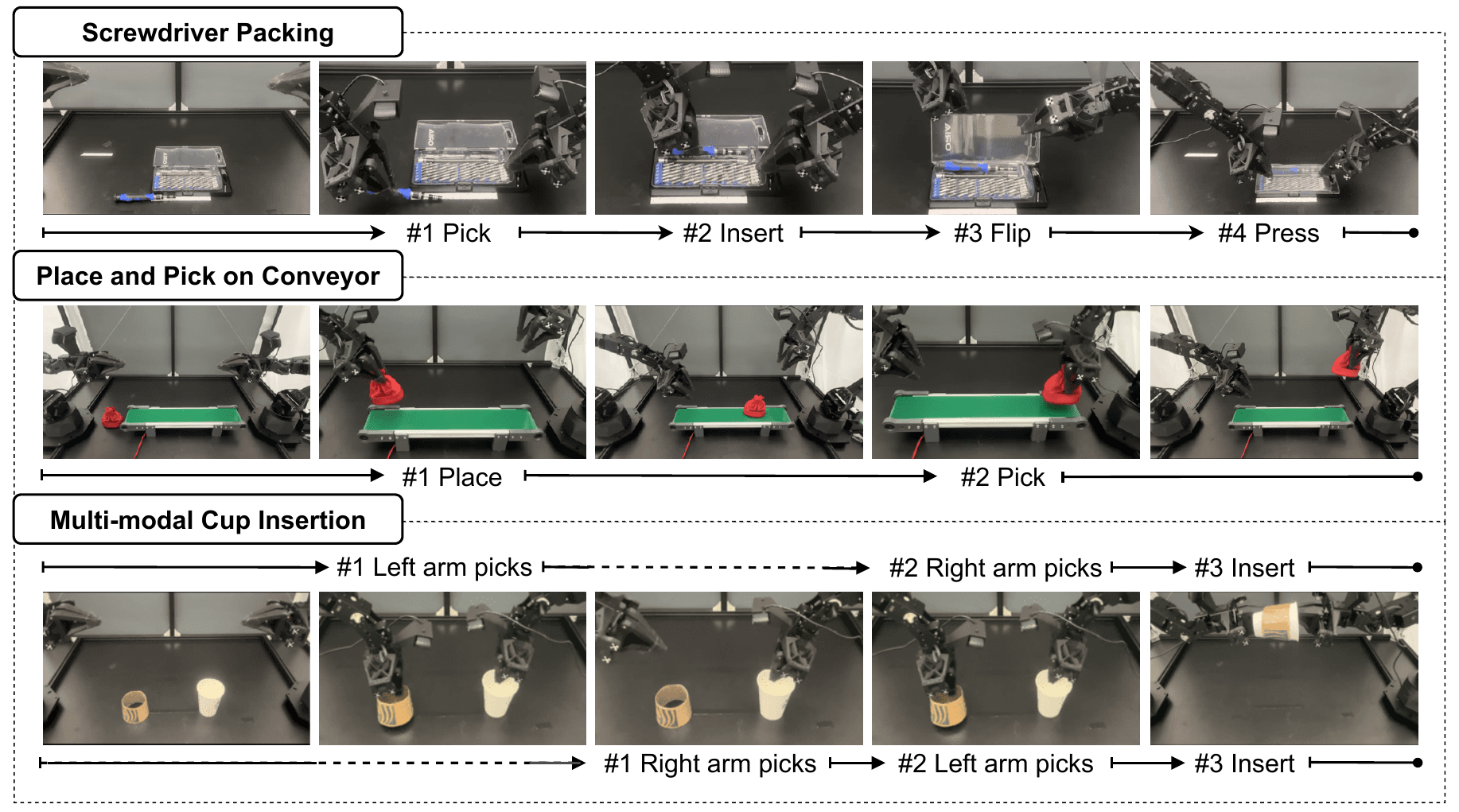}
    \vspace{-6mm}
    \caption{Processes of three real-world multi-stage bimanual tasks.}
    \label{fig:real_tasks}
    \vspace{-6mm}
\end{figure}

\noindent\textbf{Tasks and Settings.}
As shown in Figure~\ref{fig:real_tasks}, we conduct three multi-stage bimanual tasks with ALOHA, each of which focuses on a specific point.
The first task, \textbf{Screwdriver Packing}, is a long-horizon task involving multiple challenging stages, such as eye-hand coordination (e.g., picking), high precision (e.g., insertion), bimanual coordination (e.g., flipping and pressing) as well as the transparent objects (e.g., the box cover).
Due to the long horizon, this task is susceptible to the accumulation of error and latency along multiple stages, making it suitable for evaluating whether BiKC offers improved reliability and efficiency.
The second task, \textbf{Placing and Picking on Conveyor}, involves moving objects across two stages. Initially, the left arm picks a bag and places it on the conveyor belt moving at 74 mm/s. As the bag approaches the right arm via the conveyor, the right arm picks and lifts the bag when it reaches an appropriate position. 
Since the bag is moving, if the right arm's decision and execution are delayed by neural net inference, the bag may miss the suitable location, leading to task failure. 
We aim to use this task to highlight the advantages of BiKC in terms of inference speed. 
Additionally, we design a \textbf{Cup Insertion} task to insert a cup into a cup sleeve. 
We provide multi-modal demonstrations for this task: 1) The \textit{left} arm first picks the sleeve, and then the \textit{right} arm grasp the cup, and 2) The \textit{right} arm first picks the cup, and then the \textit{left} arm picks the sleeve, followed by insertion. 
We adopt this task to emphasize BiKC's capabilities of multi-modal modeling. 
All these real-world tasks involve multiple stages and bimanual coordination, making them suitable for evaluating BiKC.
Following the experimental setup outlined in previous work~\cite{zhao2023learning}, we collected 50 demonstrations for each task.
Throughout all demonstrations and rollouts, the initial workspace layout remains constant.

\begin{table}[!ht]
\centering
\caption{Success rate (\%), inference latency (ms) and total operation time (s) in three real-world tasks. All results are averaged over 20 rollouts.}
\vspace{0.1cm}
\begin{tabular}{@{}clccccc@{}}
\toprule
\textbf{Task}              & \textbf{Sub-stage\quad} & \textbf{ACT} & \textbf{DP} & \textbf{CP}  & \textbf{\textit{BiKC} (ours)} \\ 
\midrule
\multirow{7}{*}{\begin{tabular}[c]{@{}c@{}}Screwdriver \\ Packing \end{tabular}} & Pick            &100           & 100     & \multicolumn{1}{c|}{100}       & 100 \\
                          & Insert             & 50           & 30      & \multicolumn{1}{c|}{95}   & 90  \\
                          & Flip               & 100          & 50      & \multicolumn{1}{c|}{90}   & 100 \\
                          & Press              & 100          & 0       & \multicolumn{1}{c|}{18}   & 56  \\ 
                          & \textbf{Overall}     & \textbf{50.0}  & 0.0       & \multicolumn{1}{c|}{15.4}  & \textbf{50.4}    \\ 
                          \cmidrule(lr){2-6}
                          & Infer. Lat.       & 25.6        & 114.3   & \multicolumn{1}{c|}{ 22.6 }        & 26.0\\
                          & Total Dur.        & 37.0        & 33.8   & \multicolumn{1}{c|}{\textbf{24.8}}  & \textbf{25.5}\\ 
\midrule
\multirow{5}{*}{\begin{tabular}[c]{@{}c@{}} Placing and Picking \\ on Conveyor  \end{tabular}} & Put            & 95          &  100     & \multicolumn{1}{c|}{100}        &     100    \\
                          & Pick              &    0        &  10       & \multicolumn{1}{c|}{100}    &    100  \\ 
                          & \textbf{Overall}  &    0.0     &   10.0     & \multicolumn{1}{c|}{\textbf{100.0}}  &    \textbf{100.0}\\
                          \cmidrule(lr){2-6}
                          & Infer. Lat.     &    27.4    &    115.6     & \multicolumn{1}{c|}{23.3}    & 27.8     \\
                          & Total Dur.      &    29.3    &     25.5    & \multicolumn{1}{c|}{\textbf{18.4}}     & \textbf{19.7}  \\
\midrule
\multirow{6}{*}{\begin{tabular}[c]{@{}c@{}} Cup \\ Insertion  \end{tabular}}      & Pick           &    100   &  100    & \multicolumn{1}{c|}{100}       &   100 \\
                          & Insert         &   100          &  100   & \multicolumn{1}{c|}{100}         &  100 \\
                          & \textbf{Overall} &   \textbf{100}  &   \textbf{100}  & \multicolumn{1}{c|}{\textbf{100}}      &  \textbf{100} \\
                          \cmidrule(lr){2-6}
                          & Infer. Lat.    &   26.1     &   113.8  & \multicolumn{1}{c|}{23.5}       &  26.2 \\
                          & Total Dur.     &   21.5     &    20.0  & \multicolumn{1}{c|}{\textbf{14.4}}     &  \textbf{15.0} \\
\bottomrule
\end{tabular}
\label{tab:real_results}
\vspace{-3mm}
\end{table}

\begin{figure}[!t]
\centering
\subfloat[Moments when trying to pick the moving bag on the conveyor]{\label{fig:real_cm_conveyor}\includegraphics[width=\textwidth]{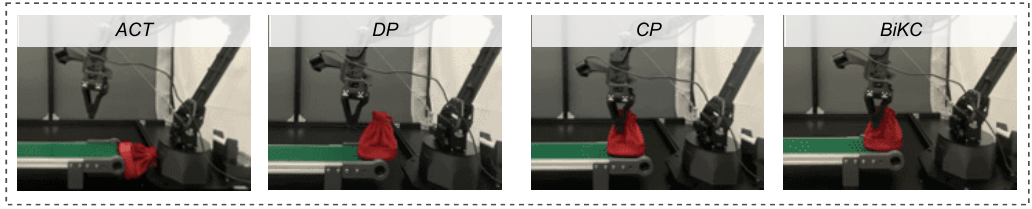}} \\
\vspace{-3mm}
\subfloat[The ``press'' of Screwdriver Packing]{\label{fig:real_cm_screwdriver}\includegraphics[width=0.5\textwidth]{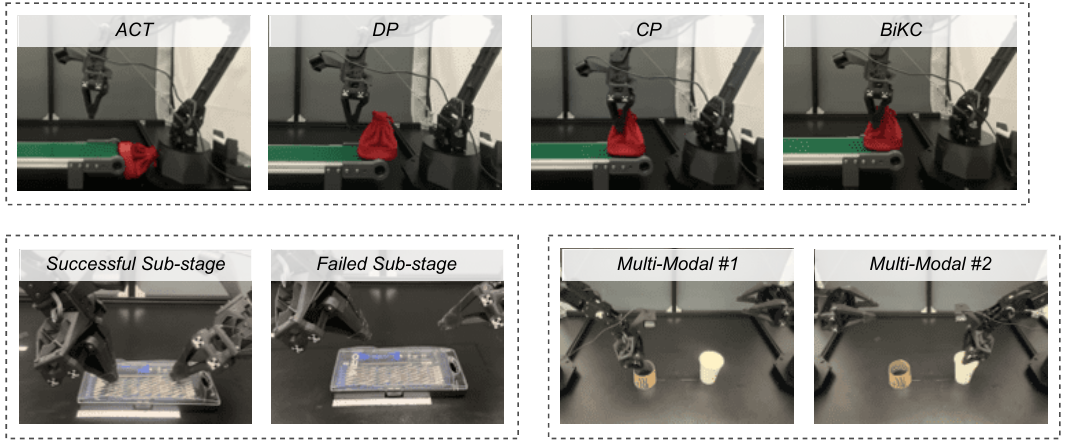}}
\subfloat[Multi-modal actions in Cup Insertion]{\label{fig:real_cm_cup}\includegraphics[width=0.5\textwidth]{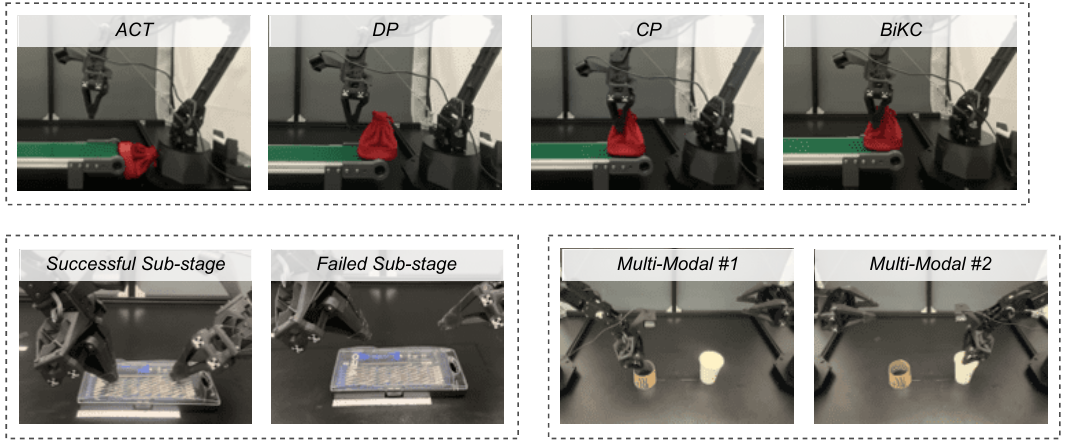}}
\vspace{-3mm}
\caption{Snapshots comparison of evaluating different algorithms in real-world.}
\label{fig:real_cm}
\vspace{-7mm}
\end{figure}

\noindent\textbf{Results and Discussion.}
We report the success rate, inference latency and total duration for three real-world experiments in Table~\ref{tab:real_results}. 
To quantify \textit{success rate} for each sub-stage task, we follow the measurement approach adopted by~\citep{fu2024mobile}. For each stage, the success rate is calculated as \textit{Success} divided by \textit{Attempts}. Here, \textit{Attempts} represents the number of successful cases from the previous stage, since the robot may fail and terminate at any stage. By employing this measurement, we can compute the final success rate for the entire task by taking the product of all sub-stage success rates.
Additionally, the \textit{inference latency} denotes the time required for the policy to output an action sequence. Operational efficiency is quantified by the \textit{total duration} for finishing a fixed number of timesteps. 
All reported results represent an average of 20 rollouts, which is a widely adopted protocol in this field of research \cite{zhao2023learning, chi2023diffusion}. Top-2 results are highlighed in Table~\ref{tab:real_results}.

In the \textbf{Screwdriver Packing} task, BiKC outperforms CP by 35\% in success rate, with the guidance from predicted keyposes. With additional condition on keyposes, BiKC is more aware of the sub-goal to be reached of short-horizon trajectories. This can be qualitatively verified by the differences in completion of the last stage of \textit{pressing}. As shown in Figure~\ref{fig:real_cm_screwdriver}, CP tends to just move the grippers down slightly and fail to close the box, while BiKC is able to finish the stage substantially.
While BiKC performs similarly as ACT in success rate and inference latency, its total operation duration (25.5s) is shorter than ACT by 11.5s, indicating superior operational efficiency.
We suppose that the temporal ensembling (TE) technique, a key feature of ACT for high success rates and smooth motion, incurs additional time costs beyond policy inference.
This experiment indicate BiKC's superior performance in terms of success rate and operational efficiency, stemming from the keypose design and CM formulation. More discussions can be found in Appendix C.

In the \textbf{Placing and Picking on Conveyor} task, both ACT and DP variants fail to stably pick up the bag due to excessive inference latency, showing lack of ability to track dynamic objects. 
In contrast, CP and BiKC achieve a success rate of 100\%, attributed to their policy formulation as consistency models, which generate action sequences by one-step inference with few latency. 
This experiment, involving dynamic factors, underscores the advantages of CM and the importance of inference efficiency as indicated in Figure~\ref{fig:real_cm_conveyor}.
Additionally, we observe that the inference latency of ACT is comparable to that of CP and BiKC. However, ACT fails to catch moving objects in a timely manner, which is attributed to TE again. Specifically, ACT predicts an action sequence at each timestep and TE performs a weighted average over historical predictions, significantly increasing decision delay. In contrast, CP and BiKC only need to perform inference every $H$ steps, enabling latency amortization and real-time reactions.

The \textbf{Cup Insertion} task involves only two stages, picking and inserting, but the first stage encompasses multi-modal actions, as depicted in Figure~\ref{fig:real_tasks}. With 25 demonstrations collected for each modality, DP and CP occasionally yet successfully generate modal \#2 action sequences (i.e., reaching cup first) during evaluation as in Figure~\ref{fig:real_cm_cup}. BiKC generates action that mixes the two modals by moving out two grippers simultaneously whereas ACT only learns to pick the sleeve first. The results highlight diffusion-based methods' capability of capturing multi-modal demonstrations, overcoming ACT's limitation.

In summary, BiKC exhibits enhancements in both success rate and operational efficiency for multi-stage bimanual tasks, outperforming state-of-the-art baselines (ACT and DP) as well as the ablated one (CP).

\section{Conclusion}
\label{sec:conclusion}

In this paper, we present a Bimanual Keypose-Conditioned Consistency Policy (BiKC) for multi-stage bimanual manipulation tasks. 
The objective of BiKC is to enhance the operational reliability and efficiency by overcoming the per-stage error and per-step latency accumulated across multiple stages.
BiKC is designed as a hierarchical imitation learning framework comprising a high-level keypose predictor that functions as a sub-goal planner, and a low-level trajectory generator that works as the behavioral policy.
The predicted keypose implies the boundary between stages, serving as guidance for action decisions and an indicator of sub-task completion.
The trajectory generator is learned from scratch as a consistency model, capable of generating an action sequence conditioned on current observation and forthcoming keypose via single-step inference.
We evaluate BiKC's performance on simulated and real-world tasks involving bimanual coordination across multiple stages. The results demonstrate BiKC's enhancements in success rates and operational efficiency, as well as its multi-modal modeling capability.

\noindent\textbf{Limitations and Future Work.}
We believe that the framework can be extended from learning and representation perspectives. Regarding \textbf{learning}, despite of accelerated trajectory generation, samples generated by CM are not as smooth as those by ACT. 
A similar phenomenon also occurs in image synthesis in the original work~\cite{song2023consistency, song2024improved}, where visual details generated by CM are compromised without high fidelity.
Although this can be mitigated by the TE trick~\cite{zhao2023learning}, the necessity of IL on short-horizon trajectories should be further verified due to the existence of modern motion planning and control approaches.
It is promising to explore whether it will be efficient and successful that IL is applied only in a coarser manner such as high-level plans and geometric constraints extracted from demonstrations.
By doing so, it becomes possible to guarantee reaching keyposes or satisfying constraints, thereby further ensuring completing sub-stages.
Regarding \textbf{representation}, we utilize joint positions as keyposes and sub-goals in this work because it aligns with the recommended state and action form of real robots, ensuring coherence in our representation. However, it is important to note that joint representations may not be as intuitive and generalized as desired, as they require forward kinematics to transform joint configurations into meaningful SE(3) poses. Furthermore, despite their popularity, we discovered that both joint positions and SE(3) poses are inherently robot-centric and may not accurately delineate the boundaries of sub-stages, as they primarily reflect the states of the robots themselves rather than capturing the task context.
The completion of manipulation tasks usually depends on desired state transition of \textit{objects} instead of robots~\cite{gao2023kvil}. Hence, it is necessary to leverage a representation that captures changes in object-centric states and further the whole workspace, and take them as sub-goals of motion. This will be more helpful for autonomous agents to understand task progress and achieve sub-goals. A scene graph is a promising formulation because it reveals the high-level relationships among objects in the workspace and can be extracted by advanced vision models~\cite{jiang2024roboexp,jiao2022sequential,zhu2021hierarchical}.

The keypose identification approach warrants further investigation. While the proposed heuristics address a range of tasks, certain situations remain out of scope. A significant challenge arises when human demonstrations are suboptimal or contain retries, a common issue in IL. In such cases, momentary pauses by human operator may be erroneously identified as keyposes, despite their irrelevance to task completion or sub-stage transitions.nother challenge emerges when sub-task boundaries lack clear definition, as exemplified by periodic tasks such as rotating a faucet handle. In such scenarios, delineating the task using a limited number of keyposes is challenging, as cyclic trajectories cannot be accurately interpolated from a finite set of intermediate steps. Future work should investigate these challenged cases in greater depth.

\begin{credits}
\subsubsection{\ackname} 
This study is partially supported by the Innovation and Technology Commission of the HKSAR Government under the InnoHK initiative as well as Research Grant Council (Ref: 17200924).

\end{credits}

\bibliographystyle{splncs04}
\bibliography{reference}

\clearpage
\begin{subappendices}
\renewcommand{\thesection}{\Alph{section}}

\section{Recapping Designs of Consistency Models}
\label{sec:cm_details}

Several improved techniques are proposed in \citep{song2024improved} to enhance performances of consistency models (CM) trained from scratch without distillation. We cover the new EMA decay rate, weighting function and loss function in Section~\ref{sec:method_details} and we recap the remaining here, including discretization, and noise schedule as well as the parameterization approach.

\noindent\textbf{Discretization curriculum} The continuous noise level $[\eps, \sigma_{\max}]$ is discretized into $N(k)$ points at the $k$-th training iteration, with $K$ iterations in total and $N(k)$ growing as $k$ increases. The new curriculum adopts a step-shaped curriculum where $N(k)$ doubles after a fixed number of iterations. Specifically,
\begin{equation}
\label{eq:discretization_curriculum}
    N(k)=\min(s_0 2^{\lfloor \frac{k}{K'}\rfloor}, s_1)+1,\quad K'=\left\lfloor \frac{K}{\log_2\lfloor s_1/s_0 \rfloor + 1} \right\rfloor,
\end{equation}
where $s_0=10,s_1=160$ are starting and end discretization numbers. Following~\citep{karras2022elucidating}, the discretized noises are given by $\sigma_i=(\eps^{1/\rho} + \frac{i-1}{N(k)-1}({\sigma_{\max}^{1/\rho}} - \eps^{1/\rho}))^\rho$, where $i\in\{1,2,\cdots,N(k)\},\rho=7,\eps=0.002,\sigma_{\max}=80$.

\noindent\textbf{Noise schedule} Although \citep{song2024improved} propose to sample $\sigma_i$ according to a lognormal distribution from the discretized sequence, we find that the original uniform distribution sampling performs better in our tasks. Therefore we stick to the old choice where $i\sim\gU\{1,2,\cdots,N(k)-1\}$.

\noindent\textbf{Parameterization} In according to satisfy the self-consistency condition mentioned in Section~\ref{sec:preliminary}, authors of CM~\citep{song2023consistency} leverage a special parameterization to make $\vf_{\bm{\theta}}(\rvx_\eps,\eps)=\rvx_\eps$ hold by design:
\begin{equation}
\vf_{\bm{\theta}}(\rvx,\sigma) \coloneqq c_{\text{skip}}(\sigma)\rvx + c_{\text{out}} \mF_{\bm{\theta}}(\rvx,\sigma),
\end{equation}
where $c_{\text{skip}}(\sigma) = {\sigma_\text{data}^2}/((\sigma-\eps)^2+\sigma_\text{data}^2)$ and $c_{\text{out}}(\sigma) = \sigma_\text{data}(\sigma-\eps)/\sqrt{\sigma_\text{data}^2+\sigma^2}$ satisfies $c_{\text{skip}}(\eps)=1$ and $c_{\text{out}}(\eps)=0$ and $\mF_{\bm{\theta}}$ is the parameterized model (e.g., a neural network).

\section{Detailed Hyperparameters}
\label{sec:hyper_params}

We generally set hyperparameters to the values suggested in the original articles for all baselines and include them in Tables~\ref{tab:hyper_ACT} and~\ref{tab:hyper_diffusion}.

\begin{table}[!ht]
\centering
\caption{Hyperparameters of ACT in real-world tasks.}
\label{tab:hyper_ACT}
\begin{tabular}{llll}
\toprule
\multicolumn{1}{c}{Parameter} & \multicolumn{1}{c}{Value} & \multicolumn{1}{c}{Parameter} & \multicolumn{1}{c}{Value} \\ \midrule
Learning rate            & 1e-5                      & \# encode layers         & 4                         \\
Batch size               & 8                         & \# decoder layers        & 7                         \\
Chunk size               & 100                       & \# heads                 & 8                         \\
KL weights               & 10                        & Feedforward dim         & 3200                      \\
\# epochs                & 10000                     & Hidden dim              & 512                       \\ \bottomrule
\end{tabular}
\vspace{-4mm}
\end{table}

\begin{table}[!htb]
\centering
\caption{Hyperparameters of Diffusion-based Algorithms on simulated and real-world tasks, where some values are in the form of [simulation | real-world].}
\label{tab:hyper_diffusion}
\begin{tabular}{@{}llll@{}}
\toprule
    Parameter & Value & Parameter & Value\\
\midrule
\textit{Shared}                      &                      & \textit{DP}                      &      \\
    \quad Learning rate              & 1e-4 $\rightarrow$ 0 & \quad \# train diffusion steps   &  100 \\
    \quad Weight decay               & 1e-6              & \quad \# eval diffusion steps       &  10 | 16 \\
    \quad Batch size                 & 64                & \quad EMA decay rate                &  0.75 \\
    \quad \# epochs                  & 500 | 1000        &     & \\
    \cmidrule(lr){3-4}
    \quad ImgRes  & 480$\times$640 | 4$\times$480$\times$640 & \textit{CM}  & \\
    \quad DownRes & 240$\times$320 | 4$\times$120$\times$160 & \quad $\sigma_{\text{data}}$ & 0.5\\
    \quad CropRes & 220$\times$300 | 4$\times$110$\times$150 & \quad EMA decay rate & 0 \\
    \quad Proprio dim                & 14 &      &  \\
    \cmidrule(lr){3-4}
    \quad Sub-trajectory horizon     & 16 &  \textit{Keypose predictor}  &  \\
    \quad Obs horizon             & 2  &  \quad Model arch  & Transformer\\
    \quad Action horizon          & 8  &  \quad Hidden dim & 256 \\
    \quad Model arch              & 1D Unet & \quad Feedforward dim  & 1024 \\
    \quad Diffusion step emb dim  & 128 & \quad \# decoder layers & 4 \\
    \quad Keypose emb dim         & 128 & \quad \# heads & 4 \\
    \quad Unet conv channel dim   & 256-512-1024 & & \\
\bottomrule
\end{tabular}

\end{table}

\section{Results Analysis of Screwdriver Packing}
\label{sec:screw_analysis}

Despite that different algorithms tend to fail at various stages (ACT and DP at ``insert'', CP and BiKC at ``press''), we suppose that they share the similar underlying cause: imitation learning performed on the entire trajectory struggles to capture fine-grained motions in demonstrations, especially when contact and force interaction are involved.
Although the overall motion tendencies can be mimicked by IL agents and the motion completion degree can be enhanced by the proposed keypose guidance, it remains challenging to reproducing subtle changes caused by contact and forces.
This is because these subtle changes are difficult to perceive solely from visual and positional sensors due to noise and sensor errors, without force-aware sensors utilized.
Therefore, without sufficient sensory information, neural networks lack the awareness of whether the screwdriver has been inserted into the box (ACT, DP), and whether the downward movements of grippers have closed the lid of the box solidly (CP, BiKC). 
The observations of whether these stages are completed or not appear highly similar in terms of images and joint positions.
To address the issue, it is necessary to involve force sensors and consider additional states such as forces and torques. 
Successful imitation of contact-rich and coordinative bimanual tasks requires not only the reproduction of motions but also the proper implementation of contact and force sequences. 
We will explore this direction as future work.
\end{subappendices}

\end{document}